\documentclass[10pt,pre,twocolumn,aps,floats,floatfix,superscriptaddress]{revtex4}

\usepackage{times}
\usepackage{helvet}
\usepackage{courier}
\usepackage{graphicx}
\usepackage{subfigure}

\begin{document}

\title{Knowledge Representation Issues in Semantic Graphs for Relationship Detection}

\author{Marc Barth\'elemy\footnote{Authors listed alphabetically.}}
\affiliation{CEA-Centre d'Etudes de Bruy\`eres-Le-Ch\^atel  \\
Departement de Physique Th\'eorique et Appliqu\'ee\\
BP12, 91680 Bruy\`{e}res-Le-Ch\^{a}tel Cedex, France \\ 
}

\author{Edmond Chow}
\affiliation{Center for Applied Scientific Computing \\ 
Lawrence Livermore National Laboratory \\ 
Box 808, L-560, Livermore, CA 94551, USA \\
}

\affiliation{Biodefense Knowledge Center,
Lawrence Livermore National Laboratory.}

\author{Tina Eliassi-Rad}
\affiliation{Center for Applied Scientific Computing \\ 
Lawrence Livermore National Laboratory \\ 
Box 808, L-560, Livermore, CA 94551, USA \\
}

\affiliation{Biodefense Knowledge Center,
Lawrence Livermore National Laboratory.}

\begin{abstract}
An important task for Homeland Security is the prediction of
threat vulnerabilities, such as through the detection of
relationships between seemingly disjoint entities. A structure
used for this task is a \emph{semantic graph},
also known as a \emph{relational data graph} or an
\emph{attributed relational graph}.  These graphs encode relationships as
{\em typed} links between a pair of {\em typed} nodes.
Indeed, semantic graphs are very similar to semantic networks used in AI.
The node and link types are related through
an \emph{ontology} graph (also known as a \emph{schema}).
Furthermore, each node has a set of attributes associated
with it (e.g., ``age'' may be an attribute of a node of type
``person''). Unfortunately, the selection of types and attributes for
both nodes and links depends on human expertise and 
is somewhat subjective and even arbitrary. This subjectiveness
introduces biases into any algorithm that operates on
semantic graphs. Here, we raise some knowledge
representation issues for semantic graphs and provide some 
possible solutions using recently developed ideas in the field
of complex networks.  In particular, we use the concept of
transitivity to evaluate the relevance of individual links in the
semantic graph for detecting relationships.
We also propose new statistical measures
for semantic graphs and illustrate these semantic measures on
graphs constructed from movies and terrorism data.
\end{abstract}

\maketitle


\section{Introduction}

A semantic graph is a network of {\em heterogeneous} nodes and links.
In contrast to the usual mathematical description of a graph, semantic
graphs have different types of nodes, and in general, different types
of links.  Also called attributed relational graphs \cite{coffman:2004}
and relational data graphs (used in the knowledge discovery literature),
it is clear that the power of these graphs lies not only in their structure
but also in the semantic information that resides on their nodes and links.
Examples of semantic graphs include citation networks where the nodes do
not simply consist of papers, but also consist of 
authors, institutions, journals,
and conferences.  Another example is the Internet Movie Database
where the nodes may be persons (actors, directors, etc.),
movies, studios, and awards, among others.  In Homeland Security,
these graphs are used in a variety of information analysis tasks
\cite{jensen.2003,coffman:2004,popp.2004,DHS-DSW:2004}.  
In particular, such graphs
may be used for predicting threat vulnerabilities.

Data for semantic graphs come from relations parsed from text documents
and/or data from relational databases.  Our motivation for this
work comes from our experience in constructing semantic graphs
from two sources of data---movies data and terrorism data---to be discussed
at the end of this paper.
In both these cases, we were faced
with a wide variety of choices:  what are the node types, what
are the link types, and how do these choices affect the algorithms
that we intend to use on these graphs?

Several types of algorithms operating on semantic graphs
are of interest to us.  For example,
to determine the nature of a possible relationship
between two entities, a subgraph consisting of the shortest paths
(or another metric) between two nodes in the semantic graph 
may be constructed and examined \cite{faloutsos:2004}.  
We refer to this process as {\em relationship detection}.
Fast algorithms based on heuristic search 
(which improve on breadth-first search or bi-directional search)
are available for this task, which either use or do not use the
semantic information in the graph \cite{eliassi-rad-tr:2004,chow-tr:2004}.  
These algorithms,
however, depend on knowing which links (or link types) in the semantic graph
are useful for detecting relationships.  For example, 
two people who share a connection to ``San Francisco'' because they
were born there are unlikely to have any real-life connection.  One of the goals
of this paper is to present automatic algorithms for determining
which are useful links for relationship detection, as well as present
concepts to help answer related questions.

In the past few years, a new field called {\em complex networks} 
(see, e.g., \citeauthor{albert:2002} (2002) and 
\citeauthor{newman:2003a} (2003)) has 
emerged to study the structure of real-world networks.  
Statistical tools for characterizing graphs and networks have been
developed, with the impetus of understanding the relationship
between the structure and function of networks.  Computer techniques have
allowed these statistical measurements to be performed on very large
real-world networks.  In this paper
we generalize some of these techniques in order to apply them
to semantic graphs.  For example, some types of nodes in semantic
graphs can be connected to many other types of nodes, but generally
have few actual links.  We quantify this concept and hypothesize that
nodes such as these are not useful for relationship detection.
In addition, the concept of {\em transitivity} in social network analysis
(called {\em clustering coefficient} in the complex networks literature)
is useful for determining
which are useful links for relationship detection.

In the following, we begin by describing semantic graphs and ontologies.
We then use the concept of transitivity for evaluating links and link
types for relationship detection.  An important aspect of this paper is
a presentation of new statistical measures for semantic graphs, as well
as issues related to the scale (level of detail) of semantic graphs.
Examples of semantic graphs for movies and terrorism data are
given near the end of the paper.

\section{Semantic Graphs and Ontologies}

A semantic graph
consists of nodes and directed links, with each
node having a {\em type} (e.g., movie).  The set of types is usually
small compared to the number of nodes.  Each node is also labeled
with one or more {\em attributes} identifying the specific node
(e.g., {\em Shrek}) or gives additional information about that node
(e.g., gross income).  Links may also have types, for example, the
(person $\rightarrow$ movie) link may be of type ``acted-in,'' or
``directed.''  (In this case, multigraphs, or graphs that may have
multiple links between the same pair of nodes, are possible.)  In some
semantic graphs, the meaning of a link between any two nodes is clear (although
different between different pairs of node types), and no link types need
to be defined.  Finally, links may also have attributes.
For additional details, see 
\citeauthor{sowa:1984} (1984).

\begin{figure}
\begin{center}
\includegraphics[width=5.0cm]{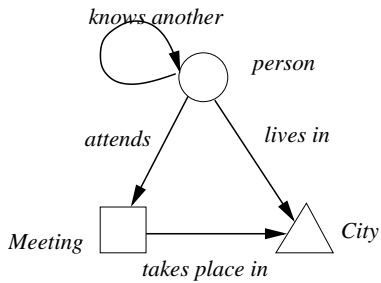}
\caption{A small ontology consisting of three node types.}
\label{fig:example}
\end{center}
\end{figure}

Depending on the types of nodes and links and on the available
information, certain relations can or cannot exist.  The set of
relations that can exist in a given semantic graph can be described by an
auxiliary graph called an {\em ontology,} 
or a {\em schema} \cite{jensen:2002}.  More often, an ontology graph 
is created first by defining the types of relations that the semantic graph
will encode.
A small example of an ontology is given in Figure \ref{fig:example},
showing three node types: person, meeting and city.  

Special links in an ontology graph could describe {\em is-a} and {\em part-of}
relationships among node types.  This is a node type hierarchy that will be
briefly mentioned when we discuss the scale of semantic graphs.

\section{Transitivity for Evaluating Nodes and Edges}

Consider a node ``San Francisco'' of type ``city'' in a semantic graph, 
and suppose we have a database of people which includes city of birth
among the data fields.  A node ``Alice'' of type ``person'' may be 
linked to the node ``San Francisco'' if Alice was born in San Francisco.
Other nodes linked to node San Francisco imply a relationship
to San Francisco and in turn their relation to Alice.  However,
it is not clear that such relationships give useful information 
about Alice since most entities a short graph distance away from ``Alice''
will have no real-life connection to Alice.  

On the other hand, people born in a city such as 
``Tikrit,'' may have a much higher likelihood of 
knowing each other, that is, it may be important in this case to be able to 
associate two people
through their city of birth.  Instead of using a human 
with potential biases to evaluate nodes
and links, an automatic procedure is
desirable for objectively determining which nodes and links
should be used in the semantic graph for relationship detection.

Another example is nodes of type ``date.''  
Dates could represent birthdates, dates of meetings, etc.
For example, a node for a person born on 9-11-2001 may be linked to a node
labeled ``9-11-2001.''
However, two events sharing a date
rarely predicts that two events are related.  Our bias is to
treat dates as attributes of nodes, rather than as its 
own node (with the type ``date'').  
Topologically, a ``date''
node may be connected to many other {\em types} of nodes, but generally
each date node is connected to only a small number of other nodes.
This may be an unbiased indication that a date is not useful for relationship
detection.

\subsection{The transitivity concept}

The concept of link transitivity is useful to address some of
the above issues.  If a node $i$ has a link to node $j$ and node $j$
has a link to node $k$, then a measure of transitivity in the network
is the probability that node $i$ has a link to node $k$.  In social
networks and many other networks categorized as {\em small-world} networks,
this probability is high.  This is natural in social networks because
a friend of a friend is also a friend in proportion that is much higher
than in a random network.  In general, we refer to $j$ as a {\em neighbor}
of $i$ if $i$ and $j$ are directly connected in a graph.  Also, we 
refer to the {\em degree} of a node as the number of neighbors it has.

The concept of transitivity is quantified as follows.
The {\em clustering coefficient} of a node, denoted by $C(i)$,
is a measure of the connectedness between the neighbors of the node.
Let $k_i$ denote the degree of node $i$, and let
$E_i$ denote the number of links between the $k_i$ neighbors.
Then, for an undirected graph, the quantity \cite{watts:1998}
\begin{equation}
C(i) = \frac{E_i}{k_i(k_i - 1)/2}
\label{eq:cc}
\end{equation}
is the ratio of the number of links between
a node's neighbors to the number of links that can exist.
We define $C(i)$ to be 0 when $k_i$ is 0 or 1.
When $C(i)$ is averaged over all nodes in the graph, we have the clustering
coefficient for a graph.
Note that
high average clustering coefficient does {\em not} imply the existence of
clusters or communities (subgraphs that are internally
more highly connected than externally) in the graph.

\subsection{Relevance of a node}

We consider the problem of determining whether a node in a semantic graph 
(e.g., ``San Francisco'' in a previous example)
is useful for relationship detection.  Consider a node $i$
which has links to many other nodes.
For now, we assume the links are of all the same type.
To evaluate whether or not $i$ is useful for relationship 
detection, we examine whether or not the neighbors of $i$ 
are actually related in the semantic graph with high frequency.
Whether or not two neighbors are related is decided by whether
or not a link exists between the two neighbors.  (A weaker condition
if this does not hold is whether the two neighbors are linked 
via a third node which is already deemed a useful node for
relationship detection.)  This leads to the use of the clustering
coefficient defined in Equation (\ref{eq:cc}) to measure 
the relevance of a node $i$ with degree greater than 1.
The equation can be generalized so that $E_i$ counts links with
the weaker condition described above.
A threshold $\tau$ is needed and if $C(i) > \tau$ then $i$ is
a useful node.  If $i$ is not a useful node, {\em all} the links 
involving $i$ should not be used for relationship detection and 
could be removed from the semantic graph.  If these links are removed,
$i$ could be made an attribute of the nodes that $i$ originally linked
to, in order not to lose any information.

The above can be generalized for semantic graphs
when $i$ is linked via many different
types of links.  In this case, instead of a count of relationships
involving pairs of neighbors of $i$, a matrix $M(t_1,t_2)$ is used
instead.  Here $M(t_1,t_2)$ counts the number of relationships
between pairs of neighbors $(a,b)$, where $a$ is linked to $i$ via type $t_1$
and $b$ is linked to $i$ via type $t_2$.  Small entries in this
matrix gives {\em pairs} of link types (associated with $i$)
that should not be traversed in relationship detection.

\subsection{Relevance of a link}

The relevance of an existing or potential relationship between two 
nodes $a$ and $b$ can be evaluated by how many neighbors they have in
common.  More precisely a relevance measure may be defined as
\begin{equation}
S(a,b) = \frac{|N(a,b)|}{|T(a,b)|}
\label{eq:strength}
\end{equation}
where
\[
N(a,b) = \left\{ w \mid w \mbox{ is linked to $a$ and $b$}, 
 w \ne a, w \ne b \right\}
\]
and
\[
T(a,b) = \left\{ w \mid w \mbox{ is linked to $a$ or $b$},
  w \ne a, w \ne b \right\}
\]
with $|T(a,b)| = \mbox{deg}(a) + \mbox{deg}(b) - |N(a,b)|$
where $\mbox{deg}(a)$ is the degree of $a$.
We have $0 \le S(a,b) \le 1$ with
large values of this relevance measure indicating a strong 
relationship between $a$ and $b$ supported by a high proportion of
common neighbors.
This quantity is similar to the clustering coefficient and
can be generalized to involve neighbors $w$ farther from $a$
and $b$.  

There are many applications of this relevance measure.  For example,
pairs of nodes with no existing link can be evaluated to check if
a latent link might exist.  In another example, the relevance measure
can be computed for all links of a given type.  A low average of this relevance
measure indicates that the given link type is not useful for 
relationship detection; there is not a strong relation between nodes
incident on a link with the given type.
A high relevance measure for a link when the average relevance measure
for the link type is low (and vice-versa)
indicates an outlier that may be interesting
to investigate.  This relevance measure must be used carefully, however,
since it uses links that it assumes confers bona fide
relationships.

It must also be recognized that a low relevance measure for an individual link
does not imply that the link is unimportant.  On the contrary, the notion
of the ``strength of weak ties'' \cite{granovetter:1973} 
suggests that these links
are critical in some sense.  It is when almost all links of the {\em same}
type have low relevance measure (and this link type is not 
$a$ ``secretly knows'' $b$) that this link type should not be used in
relationship detection.

\subsection{Generalization of clustering coefficient for semantic graphs}

\begin{figure}
\begin{center}
\includegraphics[width=3.0cm]{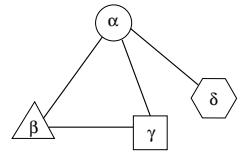}
\caption{A particular ontology for which neighbors of $\alpha$ of 
type $\delta$ can never be connected to neighbors of type $\beta$ or $\gamma$.}
\label{fig:clustering_example}
\end{center}
\end{figure}

The clustering coefficient defined earlier has little meaning for
semantic graphs as it mixes different types of nodes and it does not
include the constraints imposed by the ontology. 
To illustrate this, consider the ontology for a semantic graph
given by Figure~\ref{fig:clustering_example}.
In this case, a node of type $\alpha$ can be connected to types
$\beta$, $\gamma$ and $\delta$, but a neighbor of type $\delta$ can
never be connected to neighbors of type $\beta$ or $\gamma$. In order
to avoid unrealistically small values of the clustering coefficient we
thus have to divide by the number of links actually {\it allowed} by
the ontology and obtain
\begin{equation}
C(i;\alpha)=\frac{E_i}{E(i;\alpha)}
\end{equation}
where $E(i;\alpha)$ denotes the maximum number of links allowed
by the ontology.

\section{Statistical Measures for Semantic Graphs}

Along with clustering coefficient, two other relevant graph 
properties that have been developed for standard (non-semantic) graphs
are {\em distributions of node degree} (number of neighbors of a node)
and {\em average path length} between any two nodes in the graph.
Together, these three graph properties can be useful for
studying the properties of a semantic graph for representing knowledge.

Many real-world networks have high clustering coefficient, much higher
than $O(1/n)$ for random graphs, where $n$ is the number of nodes in
the graph.  We believe that properly constructed semantic graphs must also
have moderately high clustering coefficients.  Low values of clustering
coefficient may indicate that the linkage information in the semantic
graph is incomplete.  Very high values of clustering coefficient may
also indicate a poorly constructed semantic graph where all the nodes are
very highly linked to each other (the limit is a fully connected graph), 
indicating little discrimination in how the nodes are connected.

The average path length, $\ell$,
in a semantic graph must also not be too small (which
is also associated with very high clustering coefficients).  When the
average path length is small,
almost all nodes are approximately the same graph distance from each other,
giving little discriminatory ability to path-length based algorithms
for detecting relationships.

For example, an ontology graph may contain a node (e.g., a node of 
type ``provenance'') to which every other node in the ontology is linked.
In this case, 
the maximum shortest path length length in the ontology graph is 2,
which also suggests that the average path length in the semantic graph is
small.  It may be useful to identify nodes or links in the ontology
graph that dramatically shorten the average path length.  These nodes
and links are potentially not useful for relationship detection.

The connectivity distribution $P(k)$ 
is of interest for semantic graphs, particularly the existence of
nodes with very high degree, as in the case of scale-free
networks~\cite{barabasi:1999,amaral:2000}.  In a relationship detection
path search, paths through very high degree nodes are deemed less informative
\cite{faloutsos:2004}.  For example, 
in a social network, two people who know a popular person
are less likely to know each other; the linkages to the popular person 
should be disregarded in the relationship detection search since they
may confer erroneous relationships.

It is believed that power-law connectivity distributions arise when
there is little or no cost involved in the formation of links in the
network \cite{amaral:2000}.  Without this property, no nodes would be able to
acquire a very large number of links.  This may suggest that a graph
with power-law degree distribution may contain many weak linkages.
However, these weak linkages cannot be disregarded; Cf. strength of weak ties,
mentioned above.

For semantic graphs, we showed above how to extend the concept of
clustering coefficient.  In the next subsections, we expand the
potential usefulness of other concepts for semantic graphs.

\subsection{Extension of node degree}

\begin{figure}
\begin{center}
\includegraphics[width=5.0cm]{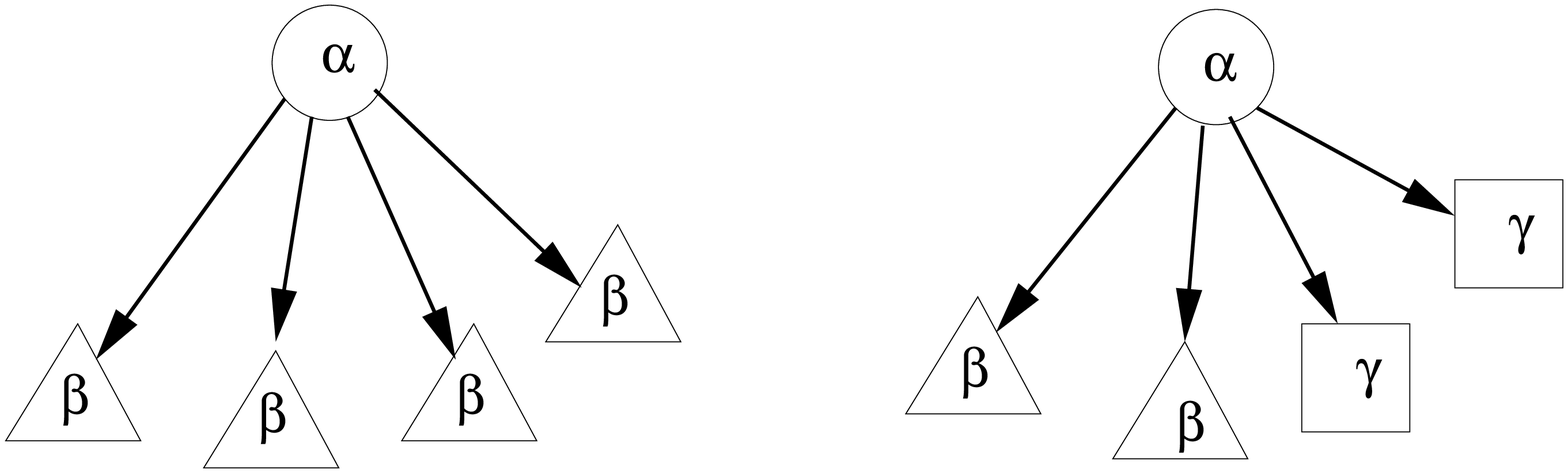}
\caption{Two examples for which the $\alpha$-type node has
topological connectivity $k=4$ but with a different meaning in each case,
Cf.~\citeauthor{jensen:2002} (2002).}
\label{fig:k_example}
\end{center}
\end{figure}

Even in the simple case of connectivity, a given value
$k$ of the connectivity of a node of type $\alpha$ has no real meaning
for semantic graphs. Indeed, as shown in Figure~\ref{fig:k_example} the
topological connectivity in both cases is $k=4$ but the meaning of it
is very different in each case.

In the first case, the environment is very homogeneous while it is not
in the second case. Another complexity comes from the
fact that the number of $\beta$-type nodes can be very large thus
inducing a bias in the connectivity of the other nodes.

The ontology implies that each node of type $\alpha$ can be connected
to a certain number, $k^{0}_{\alpha}$, of other
types. In the semantic graph, we have a total number of nodes
$n=\sum_{\alpha}n_{\alpha}$ and we denote the nodes by
$i=1,\dots,n$. The type of a node is given by the function $t(i)$. We
denote by $k_{\alpha\beta}(i)$ the number of neighbors of type $\beta$
of a node $i$ of type $\alpha$. The usual topological connectivity of
the node $i$ (which is of type $\alpha$) is then given by
\begin{equation}
k_{\alpha}(i)=\sum_{\beta}k_{\alpha\beta}(i).
\end{equation}
Using this quantity, we can define the average connectivity
of type $\alpha$ which is just the average over all nodes with
type $\alpha$ as
\begin{equation}
\overline{k_{\alpha}}=\frac{1}{n_{\alpha}}\sum_{i,\; t(i)=\alpha}k_{\alpha}(i).
\end{equation}

If we want to compare the different types relative to their
connectivity, it is important to remember that some types can be
connected to many others (such as persons which can be linked to
others persons, cities, meeting, jobs, etc.) while other types are
only linked to one type (such as a conference which takes place only at
one location). In order to compare the different types we thus have to
rescale by the number of different neighbor types they can have according to
the ontology:
\begin{equation}
m_{\alpha}=\frac{\overline{k_{\alpha}}}{k^{0}_{\alpha}}.
\end{equation}

This quantity indicates the average number of neighbors {\it per
type}. This quantity however does not tell us if there are large
connectivity fluctuations or if in contrast all nodes of a given type
have essentially the same connectivity. We thus have to measure the
connectivity variance {\it per type} which is calculated using the second moment
\begin{equation}
\overline{k^{2}_{\alpha}}=\frac{1}{n_{\alpha}}\sum_{i,\; t(i)=\alpha}k^{2}_{\alpha}(i)
\end{equation}
with the dispersion per type given by
\begin{equation}
\sigma^{k}_{\alpha}=\frac{[\overline{k^{2}_{\alpha}}-(\overline{k_{\alpha}})^2]^{1/2}}
{k^{0}_{\alpha}}.
\end{equation}

Another possible way to characterize the connectivity distribution per
type is to plot the connectivity distribution. However, the dispersion
around the average is already a first indication of the nature of the
connections for different types. For some cases, the fluctuations
will be small, while for others it can fluctuate greatly 
(such as the number of persons a person knows).

\subsection{Disparity of connected types}

The above quantities tell us the expected number of connections of a node of
a given type to another type
but not the correlations between different types. Indeed, a type
$\alpha$ can preferentially link to a type $\beta$ while it could be
in principle also be linked to other types (as given by the ontology).

We thus quantify the disparity (or affinity) of each
type to link to other types. In order to do this we use a convenient
quantity---denoted by $Y_2$---which was introduced in another
context~\cite{Derrida:1987,Barthelemy:2003a}. In order to understand the meaning
of this quantity let us consider an object that is broken into a number $N$
of parts, each part having a weight $w_i$. By construction $\sum_{i}w_i=1$
and $Y_2$ is given in this case by
\begin{equation}
Y_2=\sum_i[w_i]^2.
\end{equation}
If all parts have the same weight $w_I\sim 1/N$ then $Y_2\sim 1/N$ is
small (for large $N$). In contrast, if we have $w_1=1/2$ and the rest
is small implying $w_{i\ne 1}\sim 1/2(N-1)$ then we obtain $Y_2\sim
1/4$. This simple example can be easily generalized to more
complicated situations and shows that a small value of $Y_2$ indicates
a large number of relevant parts while a larger value (typically of
order $1/m$ where $m$ is of order unity) indicates the dominance of a
few parts.

We now apply this idea to the number of types to quantify the
disparity of a node or the affinity of a type. The quantity $Y_2$ is first
defined for a given node $i$ of type $\alpha$
\begin{equation}
Y_2(i;\alpha)=\sum_{\beta}\left[\frac{k_{\alpha\beta}(i)}{k_{\alpha}(i)}\right]^2.
\end{equation}
In order to get results with statistical significance, we average this 
quantity over all
nodes of the same type and we also compute its dispersion $\sigma^{Y}_{\alpha}$:
\begin{eqnarray}
\overline{Y}_2(\alpha)=\frac{1}{n_{\alpha}}\sum_{i,\; t(i)=\alpha}
Y_2(i;\alpha),\\
\sigma^{Y}_{\alpha}=\left[
\overline{Y_2^{2}(\alpha)}-(\overline{Y}_2(\alpha))^2
\right]^{1/2}.
\end{eqnarray}

These results must however be weighted by the fact that some types are
more numerous than others which could be a reason why they appear
more often than others. For a given node $\alpha$, we denote by ${\cal
V}(\alpha)$ the set of types which can be connected to $\alpha$ as
given by the ontology. If a node has $k$ neighbors, and if these
neighbors are picked at random in the set of different nodes with
population $n_{\beta}$, we then obtain a disparity given by
\begin{equation}
Y^{r}_2=\sum_{\beta\in{\cal V}(\alpha)}\left[\frac{n_{\beta}}{n}\right]^2.
\end{equation}
Again, this quantity will be very small if all types are uniformly
present in the semantic graph $Y^{r}_2\sim 1/N$ (where $N$ is the
total number of different types) and if it is of order unity then
essentially a few types are over-represented. In order to take these
heterogeneities into account it is thus necessary to rescale
$Y_2(\alpha)$ by $Y^{r}_2$ and to form the factor
\begin{equation}
R(\alpha)=\frac{Y_2(\alpha)}{Y^{r}_2}
\end{equation}
and its corresponding dispersion,
\begin{equation}
\sigma^{R}_{\alpha}=\frac{\sigma^{Y}_{\alpha}}{Y^{r}_2}.
\end{equation}

A large value (larger than one) of $R(\alpha)$ indicates that type
$\alpha$ preferentially links to a small number of types and that 
its neighbor types ${\cal V}(\alpha)$ are diverse in number.
If $R\ll 1$, the type
$\alpha$ may still be preferentially connected to a small set of types
but the diversity of the numbers of each neighbor type is small.

The dispersion $\sigma^{R}(\alpha)$ indicates whether the behavior as
described by the average value $R(\alpha)$ is typical, or if in
contrast there is large diversity among the nodes of type $\alpha$.

Other usual quantities that are measured in order to 
characterize a large network can also be generalized without any
difficulty. For example, degree distributions should be examined by
type of node.  In a semantic graph, the overall degree distribution
may not be meaningful, but the degree distribution for a specific
node type may be power-law, etc.
As a further example, the average path length generalizes to become a matrix
$\ell_{\alpha\beta}$ where $\alpha$ indicates the source node of the
shortest paths while $\beta$ is the target node. 
This matrix will in general have
entries with very different values.

%

\section{Scale in Semantic Graphs}

Given a knowledge base of relational data, the choice of ontology
depends on what information needs to be captured in the semantic graph,
and how easily certain information needs to be retrieved.
The level of detail (or scale) chosen for the ontology
(choice of node and link types)
will have a direct impact on the properties of the corresponding
semantic graph. 

In the simplest ontology, we have nodes of only one type.
In the example of the movies database, this ontology
is a simple network of actors without any types and two actors are
connected if they played in the same movie.  At the next finer
scale, we have actors and movies as node types.  In this
case, the ontology is an actor connected to a movie if he played in that
movie.  This is a special case of a semantic graph
which is a {\em bipartite} network (two types of nodes, with links only between
the two types).
Coarser models lose some of the information present in finer models
but can be useful for
large-scale computations, such as multi-level search techniques.

At the finest scale of a terrorist network, we may have nodes
of type ``Religious Terrorist Organization'' and ``Political Terrorist 
Organization.''  A coarser model may aggregate nodes of these
two types into a new type, ``Terrorist Organization'' (or the
aggregation may occur directly if a type hierarchy is available).
Depending on what information needs to be preserved, it may or may not be
important to distinguish between these two node types
at the structural level of the semantic graph.

We note that in Homeland Security tasks, 
data analysis more often involves searching for outliers rather 
than commonplace patterns.  Thus it is essential that the fine
scale data is retained and the coarse scale data is used
appropriately (for example, as an aid in managing and processing
large-scale data).

%
%
%
%


\subsection{Effect of scale on statistical measures}

Here we simply illustrate the effect of scale on the clustering coefficient.
We consider a random bipartite graph with Poisson distributed
numbers of both movies per actor (with average $\mu$) and actors per
movie (with average $\nu$). We suppose that we have $n_A$ actors and
$n_M$ movies and the fact that each link connects an actor to a movie imposes
the constraint
\begin{equation}
\frac{\mu}{n_A}=\frac{\nu}{n_M} .
\end{equation}

This model can be considered as a ``null'' model since there are no
particular correlations here. If one computes the clustering coefficient of
the one-mode projection
of this network, one obtains~\cite{newman:2001a}
\begin{equation}
C=\frac{1}{\mu+1} .
\end{equation}
This quantity is finite even in the limit of very large networks
$n_{A,M}\to\infty$. This is in contrast with the usual random network for
which
\begin{equation}
C\sim \frac{1}{n}
\end{equation}
where $n$ is the number of nodes. At this stage the conclusion is that
the actor network is very clustered and different from a random
network with no correlations. This is however clearly an incorrect
statement since the existence of a large clustering coefficient here is a
consequence of the network construction procedure.


\section{Examples}

\subsection{Movies data}   
                                       
\begin{figure}
\begin{center}
\includegraphics[width=2in]{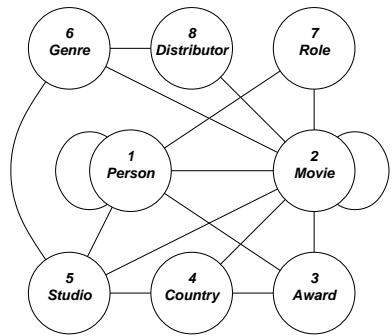}
\caption{Movies ontology.}
\label{fig:imdb-ont}
\end{center}
\end{figure}

The ``Movies'' test data at the UCI KDD Archive contains information
about movies, persons (actors, directors, etc.), studios, awards, etc.
The data was originally compiled by Gio Wiederhold (Stanford University).
We used this data to construct an ontology and semantic graph to 
express most of the information in the dataset.  Figure \ref{fig:imdb-ont}
shows the ontology graph that we developed.  In the figure, the meaning of
most of the links is obvious.  However, the person-person 
link implies {\em married-to}, {\em lived-with}, or some other non-professional
relationship; the person-studio link implies {\em founded}; the movie-movie
link implies {\em sequel-to}.  We note that the data is very incomplete.

In this ontology, the best meaning of the node Role is unclear.
For example, are two actors linked to the same Role node in the semantic
graph if they
played the role of Villain in two different movies?  Alternatively,
a role node in the semantic graph may only link to actors playing
a given role in a single movie.  We arbitrarily chose the former in our case.

A related question, which is structurally similar but semantically different
is the following.  Should two actors who win a Best Actor award be linked to the
{\em same} Award node in the semantic graph?  In this case we did not choose
this interpretation since it seems that awards are individual entities,
whereas roles are not.

Table \ref{tbl:imdb-results} summarizes the node types, frequencies,
and other statistical measures for the movies semantic graph.
The results show
high dispersion of average connectivity per type, for all types.
Further, the disparity of connected types is not particularly 
different from a random model.  These indicate a relatively well-constructed
semantic graph; there are no particular correlations (given
the numbers of each node type) and thus the information
content in the graph is high.  The results will be very different
for the terrorism data.

In the semantic graph,
the nodes with the largest clustering coefficients 
depend on whether the types of the nodes are considered.  In the standard
case where the types are not considered, the node Maurice Barrymore
has high clustering coefficient; the node is connected to Georgiana Drew
Barrymore, Lionel Barrymore, Ethel Barrymore, etc., all of which
are connected to each other.  If node types are considered, then it is 
not important that neighbors of a node are not linked if they are 
not permitted to be linked according to the ontology.  Now nodes that
were missed with the above measure may have high clustering coefficient,
e.g., the movie {\em Dogma} (perhaps due to the idiosyncrasies of 
the incomplete data).


In the semantic graph, the link between Columbia Pictures and 
drama (genre) has the most number of common neighbors (710).  
However, when the link
relevance measure (Equation (\ref{eq:strength})) is used,
which accounts for the number of links a node has,
the link between Bud Abbott and Lou Costello is found (30 common neighbors).
(We also found re-releases of movies under a new name in this process.)
Further, a semantic version of relevance can be defined, which
considers only the links that are allowed by the semantic graph.
In this case, the link between Tokuma Studio and docu-drama is found.
(Tokuma is linked to drama and the movie {\em Carences}; docu-drama is 
linked to {\em Carences} and Miramax; and Miramax is linked to drama.)

We also computed the average relevance per link type for the semantic graph.
First, the link types of least frequency were Person-\emph{founded}-Studio
and Studio-\emph{located-in}-Country.  However, the links with lowest average
relevance per link were Movie-\emph{shot-in}-Country and 
Award-\emph{awarded-in}-Country.  As mentioned, these latter links may by
least useful for automatic relationship detection.

\begin{table}
\begin{center} \scriptsize
\begin{tabular}{|rl|r|rr|rr|} \hline
   & Node Type       & $n_\alpha$ & 
        $m_\alpha$ & $\sigma_\alpha^k$ & $R(\alpha)$ & $\sigma_\alpha^R$ \\
\hline
 1 & Person          & 21504  &    0.872 &    2.383 &   1.836 &    0.663 \\
 2 & Movie           & 11540  &    1.131 &    0.816 &   1.299 &    0.644 \\
 3 & Award           &  6734  &    2.579 &   10.201 &   0.905 &    0.144 \\
 4 & Country         &    19  &  222.509 &  582.572 &   1.812 &    0.364 \\
 5 & Studio          &  1075  &    1.948 &    9.534 &   1.241 &    0.408 \\
 6 & Genre           &    39  &   77.803 &  160.060 &   0.512 &    0.154 \\
 7 & Role            &   115  &   25.561 &   64.164 &   0.924 &    0.028 \\
 8 & Distributor     &    16  &  206.156 &  356.043 &   0.782 &    0.165 \\
\hline
\end{tabular}
\caption{Node types and statistics for the movies data:  frequency of
node type $n_\alpha$, average connectivity per type $m_\alpha$ and 
its dispersion $\sigma_\alpha^k$, disparity of connected types
$R(\alpha)$ and its dispersion $\sigma_\alpha^R$.  
The results show
high dispersion of average connectivity per type, for all types.
Further, the disparity of connected types is not particularly 
different from a random model.
}
\label{tbl:imdb-results}
\end{center}
\end{table}

\subsection{Terrorism data}

\begin{figure}                         
\begin{center}                         
\includegraphics[width=2in]{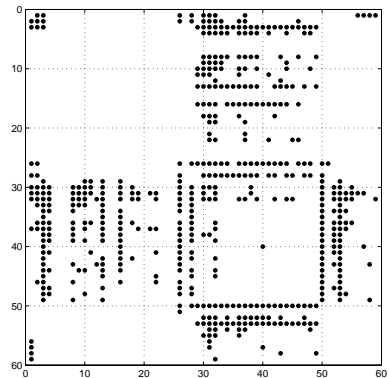}      
\caption{Adjacency matrix for the terrorism ontology.  The matrix
is used to determine which node types are allowed to link to a given type.}
\label{fig:terr-adjmat}                
\end{center}                           
\end{figure}

Relational data about world-wide terrorist events is available,%
\footnote{Data available at http://ontology.teknowledge.com.}
as well
as ontologies describing the organization of this data \cite{niles:2001}.
From this data we constructed an ontology and semantic graph.
The 59 node types are shown in Table \ref{tbl:terrorism-types}.
The ontology is shown in Figure \ref{fig:terr-adjmat} as an adjacency
matrix.  The semantic graph contains 2366 nodes.


\begin{table}
\renewcommand{\arraystretch}{0.6}
\begin{center} \scriptsize
\begin{tabular}{|rl|c||rl|c|} \hline
   & Type & $n_\alpha$ & & Type & $n_\alpha$ \rule[-1.0ex]{0pt}{3ex}\\
\hline
 1 & Nation                       &  92 & 31 & Shooting               & 445 \\
 2 & GeographicalRegion           &  85 & 32 & Bombing                & 323 \\
 3 & City                         & 555 & 33 & HostageTaking          &  14 \\
 4 & Building                     &  10 & 34 & IncendDeviceAttack     &  18 \\
 5 & Combustion                   &   0 & 35 & Lynching               &   3 \\
 6 & Destruction                  &   0 & 36 & SuicideBombing         & 107 \\
 7 & Device                       &   0 & 37 & CarBombing             & 114 \\
 8 & GeographicArea               &   3 & 38 & Arson                  &  15 \\
 9 & Government                   &   1 & 39 & HandgrenadeAttack      &  38 \\
10 & GovernmentPerson             &   2 & 40 & Hijacking              &  15 \\
11 & Group                        &   1 & 41 & RocketMissileAttack    &  14 \\
12 & Hole                         &   1 & 42 & KnifeAttack            &  53 \\
13 & Human                        &   6 & 43 & ChemicalAttack         &   9 \\
14 & JoiningAnOrg                 &   0 & 44 & LetterBombAttack       &  10 \\
15 & Killing                      &   0 & 45 & Stoning                &   3 \\
16 & OccupationalRole             &   3 & 46 & VehicleAttack          &   7 \\
17 & Region                       &   0 & 47 & MortarAttack           &   8 \\
18 & SocialRole                   &   1 & 48 & Vandalism              &   4 \\
19 & StationaryArtifact           &   1 & 49 & Other                  &   5 \\
20 & UnilateralGetting            &   0 & 50 & Number                 & 120 \\
21 & Vehicle                      &   1 & 51 & Continent              &   2 \\
22 & ViolentContest               &   1 & 52 & GeneralStructure       &   6 \\
23 & Weapon                       &   0 & 53 & Month                  &  12 \\
24 & Proposition                  &   0 & 54 & GeneralBuilding        &   2 \\
25 & BinaryPredicate              &   0 & 55 & GeneralHuman           &   2 \\
26 & ForeignTerrOrg               &  28 & 56 & Airbase                &   2 \\
27 & ReligiousOrg                 &   0 & 57 & Airport                &   3 \\
28 & TerroristOrg                 &  53 & 58 & State                  &   4 \\
29 & Infiltration                 &   8 & 59 & Railway                &   1 \\
30 & Kidnapping                   & 155 &    &                        &     \\
\hline                                 
\end{tabular}                          
\caption{Node types and their frequencies, $n_\alpha$, for the terrorism data.}
\label{tbl:terrorism-types}            
\end{center}                           
\end{table}                            
                                       
\begin{figure}                         
\begin{center}                         
\includegraphics[width=3.25in]{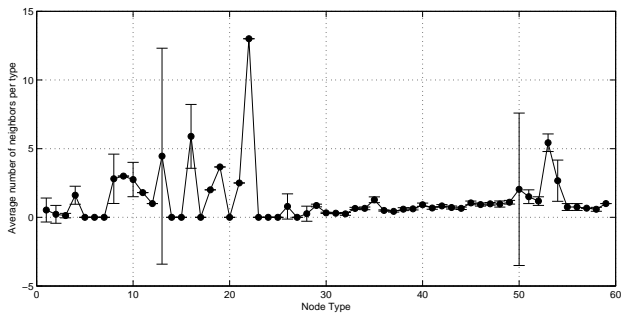}
\caption{Terrorism data: average number of neighbors per type, $m_\alpha$.  
Each error bar is of 
length $\sigma_\alpha^k$ on each side of the average.  }
\label{fig:terr-deg}                
\end{center}                           
\end{figure}                           

\begin{figure}                         
\begin{center}                         
\includegraphics[width=3.25in]{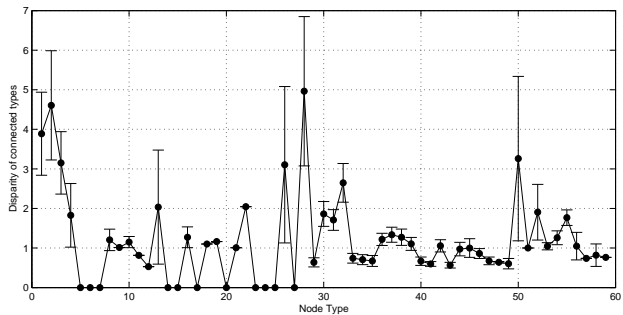}
\caption{Terrorism data: 
disparity of connected types, $R(\alpha)$.  Each error bar is of 
length $\sigma_\alpha^R$ on each side.}
\label{fig:terr-con}                
\end{center}                           
\end{figure}                           

Figures \ref{fig:terr-deg} and \ref{fig:terr-con} plot the average number
of neighbors per type and the disparity of connected types, respectively.
Error bars are used to show the dispersion of the quantities.
We consider that frequencies of 50 or more in this data set are 
statistically significant.  Thus, we consider types
1, 2, 3, 28, 30, 31, 32, 36, 37 42, and 50.
For all these types, the average number of neighbors per type is small.
The types, however, can be separated by their disparity.
Types 1, 2, 3, 28, and 50 have high disparity, i.e., they are connected
to many different types.  This is consistent with nodes of
types 1, 2, and 3 being of type 
``location,'' nodes of type 28 being of type ``terrorist organization,''
and nodes of type 50 being of type ``number.''
The remaining types are types of attacks and are not particularly
correlated with any other node types (given the numbers of each node type).
We note in this case
that semantically similar node types
have similar values of $m_\alpha$ and $R(\alpha)$.

\section{Conclusion}

This paper reveals some of the knowledge representation
issues associated with semantic graphs.  Ideas from the field of complex
networks have been applied and generalized to semantic graphs.
For example, transitivity may be used to determine
the relevance of edge types for relationship detection.

We have defined several measures for statistically characterizing
node types.  These quantities
take into account the ontology which specifies the permitted connections in
the semantic graph.
Many other important measures can be defined,
such as correlations with attribute {\em values}
\cite{jensen:2002}, which was not covered in this paper.
These and other tools can be useful to help design ontologies
and semantic graphs for knowledge representation.

\section{Acknowledgments}
We are pleased to
thank Keith Henderson and David Jensen for helpful discussions.
MB wishes to thank the Center for Applied Scientific Computing and
the Institute for Scientific Computing Research at Lawrence Livermore
National Laboratory for their hospitality during the formative stages
of this work. This work was performed under the auspices of the U.S. Department
of Energy by University of California Lawrence Livermore
National Laboratory under contract No.~W-7405-ENG-48.


\end{document}